\ifx\XeTeXrevision\undefined
  \pdfoutput=1
\fi
\documentclass[10pt]{article}

\usepackage{LobsterTwo}
\usepackage{xspace}

\usepackage{preprint}
\colorlet{ColaBlue}{accent!90!black}

\colorlet{HeaderBlue}{accent!13}

\colorlet{ColaTint}{accent!7}

\colorlet{GroupGray}{accent!5}

\colorlet{DropGray}{accent!75!black}

\newcommand{\colaGainUp}[1]{%
  {\fontsize{6.8}{7.8}\selectfont
   \color{ColaBlue}$\uparrow$\,#1\%}%
}

\newcommand{\colaBestGain}[2]{%
  \makecell[c]{%
    \textbf{#1}\\[-1pt]
    \colaGainUp{#2}%
  }%
}

\colorlet{EffHeaderBlue}{accent!12}

\colorlet{EffColaTint}{accent!6}

\colorlet{EffInkBlue}{accent!90!black}
\usepackage{booktabs}
\usepackage{makecell}
\usepackage{array}
\usepackage{colortbl}

\newcommand{\effBestDelta}[2]{%
  \shortstack[c]{%
    \textbf{#1}\\[1.5pt]
    {\fontsize{7}{8}\selectfont
     \color{EffInkBlue}$\downarrow$\,#2\%}%
  }%
}

\newcommand{\effDelta}[2]{%
  \shortstack[c]{%
    #1\\[1.5pt]
    {\fontsize{7}{8}\selectfont
     \color{EffInkBlue}$\downarrow$\,#2\%}%
  }%
}
\usepackage{microtype}
\usepackage{amsmath,amssymb}
\usepackage{algorithm}
\usepackage{algpseudocode}
\usepackage{multirow}
\usepackage{float}
\usepackage{tikz}
\usetikzlibrary{arrows.meta}
\usepackage{hyperref}
\usepackage[capitalise,noabbrev,nameinlink]{cleveref}

\DeclareRobustCommand{\name}{%
    {\LobsterTwo Coca-Cola}\xspace
}

\logos{%
  \mbox{%
    \raisebox{-0.5\height}{%
      \includegraphics[height=0.30in,
        trim=465bp 80bp 490bp 75bp,clip]{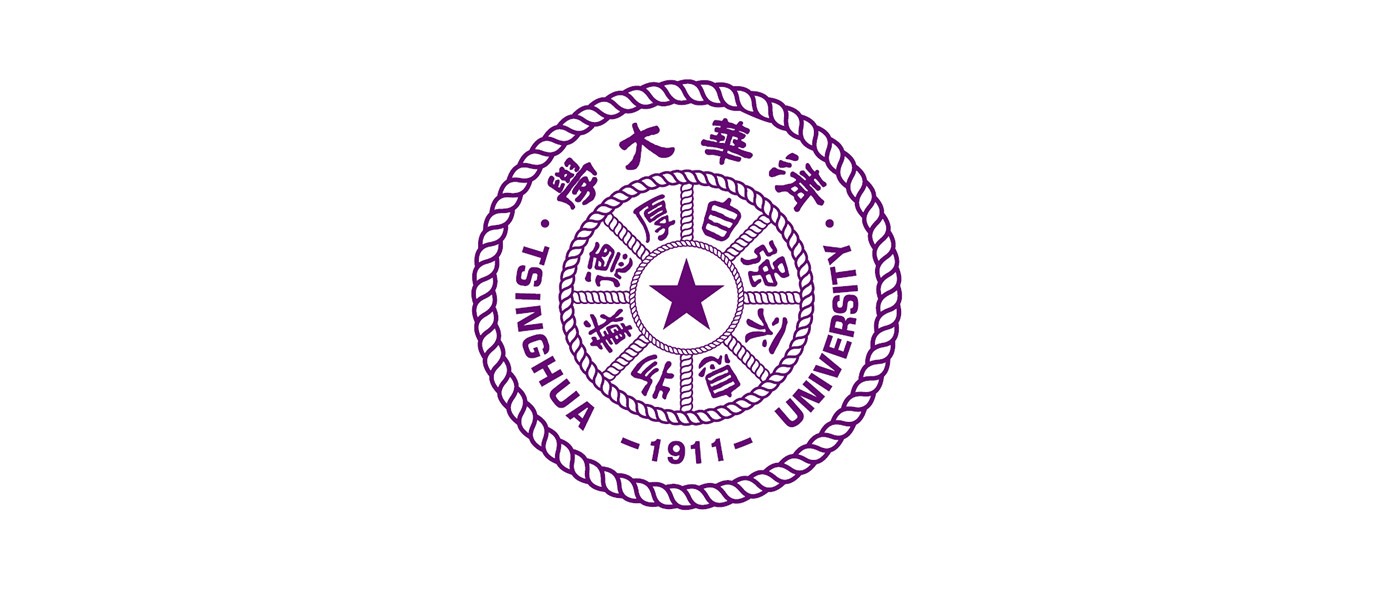}}%
    \hspace{7pt}%
    \raisebox{-0.5\height}{%
      \includegraphics[height=0.32in,
        trim=0bp 0bp 0bp 120bp,clip]{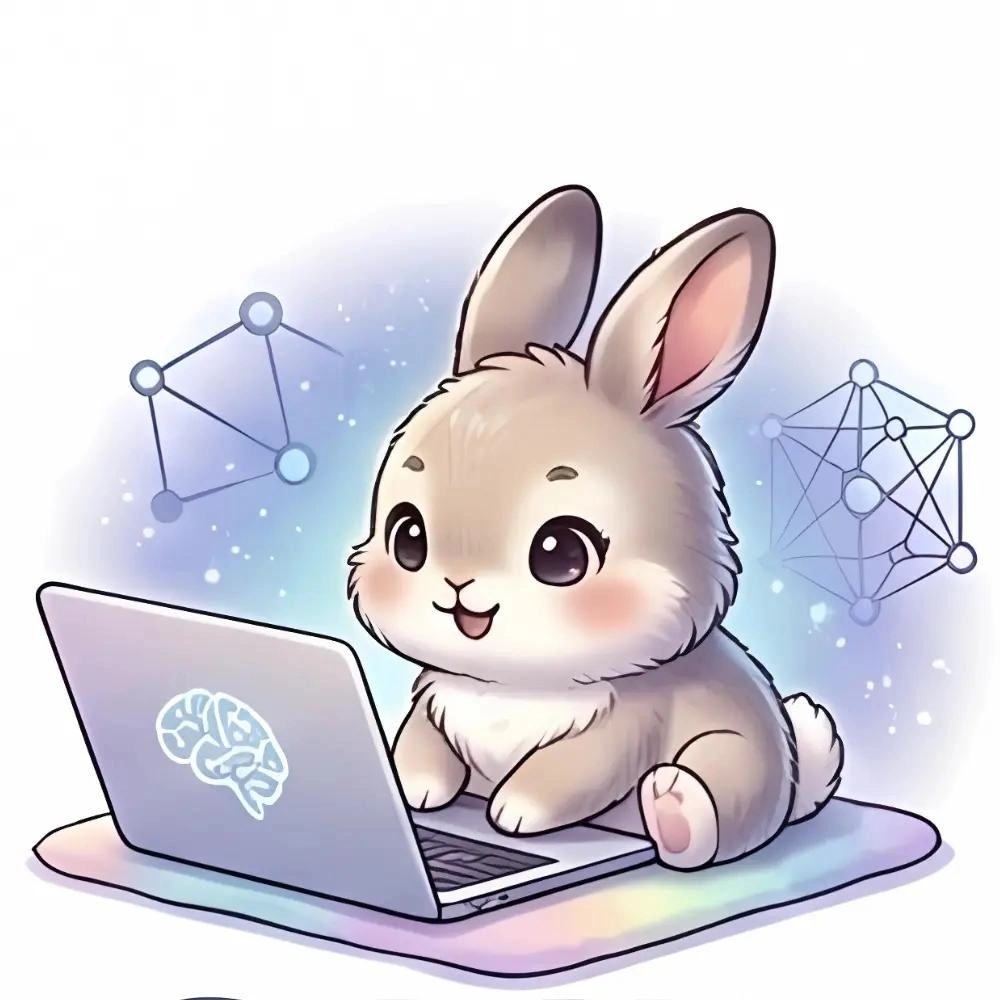}}%
    \hspace{7pt}%
    \raisebox{-0.5\height}{%
      \fontfamily{Montserrat-LF}\fontseries{medium}\fontshape{n}%
      \fontsize{10.5}{12}\selectfont\color{accent}%
      \ifPDFTeX\textls[10]{\groupname}\else\groupname\fi}%
  }%
}

\title{Tools Need Not Speak in Text:\\ Latent Context Compression for Agents}
\author[1]{Yuetong Liu}
\author[1]{Zhehao Zhou}
\author[1]{Yitong Zhang}
\author[1]{Haorui Chen}
\author[1,\dagger]{Jia Li}
\affiliation[1]{\groupname, Tsinghua University}
\date{\today}
\metadata[First author]{\email{23371522@buaa.edu.cn}}
\metadata[Corresponding author]{$^{\dagger}$\,\email{jia\_li@mail.tsinghua.edu.cn}}

\begin{document}
\maketitle

\begin{abstract}

Large language model (LLM) agents typically append raw tool
outputs to their context. Over multi-step interactions, long
and noisy observations incur substantial context overhead and
dilute the agent’s attention to task-relevant information. Existing
token pruning and summarization methods remain constrained by discrete text representations, limiting the achievable compression
ratios.

In this work, we challenge the prevailing assumption that agents must consume tool observations as explicit text tokens. To this end, we propose \name, a continuous latent context compression model for LLM agents. \name compresses tool observations into latent tokens and embeds them into the input embedding space of the agent. We further introduce \textbf{Action-Preserving Training}, which aligns the agent's next-action distributions under full and compressed observations. This trains the compression model to learn
action-sufficient latent tokens that retain decision-critical information. Experiments with web-search agents on four QA benchmarks and two model families show that \name reduces the agent's token consumption by 70.01\%--89.24\% and end-to-end runtime by 24.45\%--55.97\%, while consistently improving task performance. These results suggest that \name offers a promising approach for context compression in agents. Our code is available at \url{https://github.com/THU-Agent/Coca-cola}.
\end{abstract}

\section{Introduction}
\label{sec/intro}

Large language model (LLM) agents increasingly use
external tools to acquire information and solve
multi-step tasks~\cite{sweagent,data_interpreter,voyager}, with web search as a representative
example~\cite{toolformer,search_o1,webvoyager,webthinker}. Under the widely adopted ReAct~\cite{react} paradigm, an agent first reasons over the task and its current context to select a tool call, then adds the returned observation to its context to take the next action. For example, after querying a search engine, a web-search agent appends the returned search results to its context and uses them to decide whether to perform another search or provide a final answer~\cite{search_o1}.

\begin{figure*}[t]
    \centering
    \includegraphics[width=\textwidth]{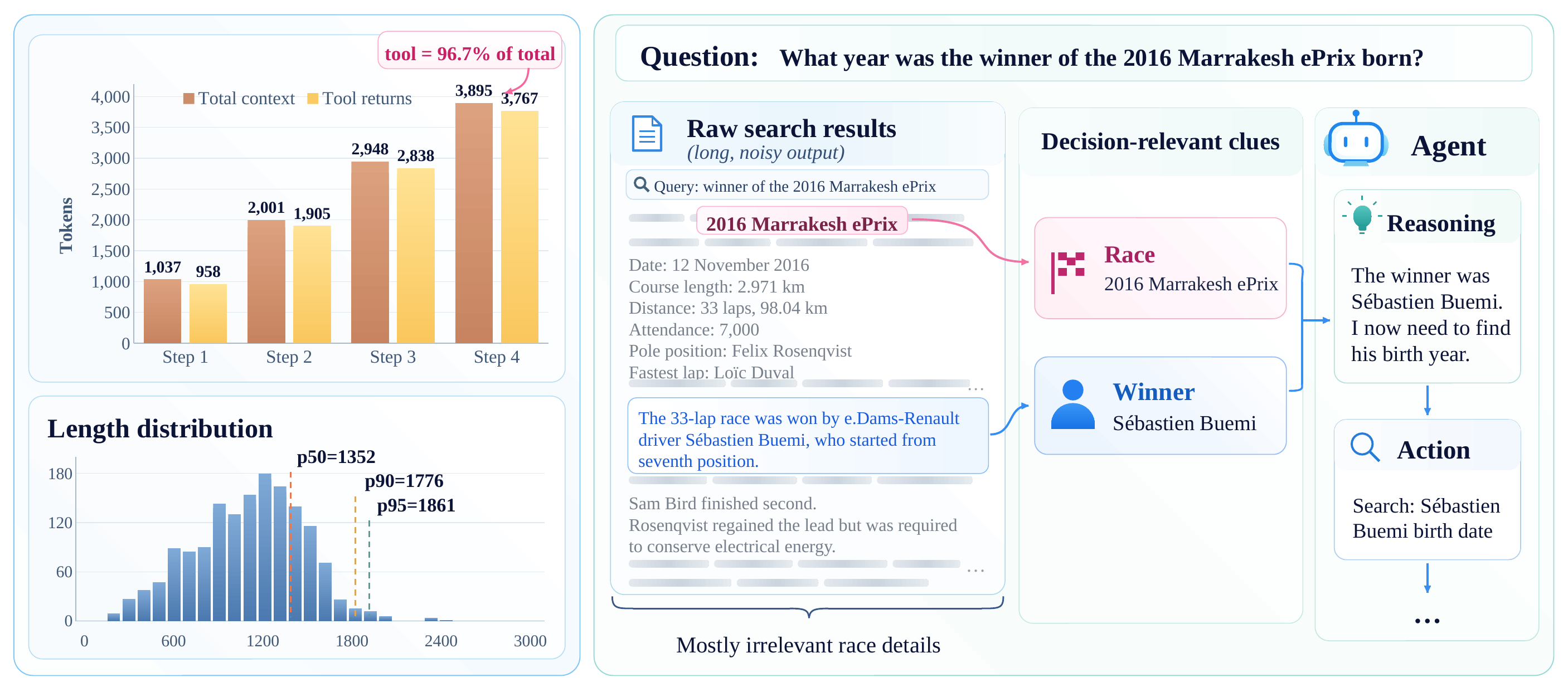}
    \caption{\textbf{Tool observation redundancy in LLM agents.}
    \textbf{Left:} Tool returns increasingly dominate the agent context, accounting for 96.7\% of total tokens by Step 4, while individual observations exhibit substantial length.
    \textbf{Right:} A representative web-search example illustrating that only a few fragments of a lengthy tool output are relevant to the agent’s next decision.}
    \label{fig:intro}
\end{figure*}

However, appending raw tool observations leads to rapid context growth over successive reasoning steps, introducing two drawbacks. \textbf{First, increased computational overhead.} As shown in Figure~\ref{fig:intro} (left), tool observations account for a large and growing share of the context, reaching 96.7\% at Step~4 and leading to substantially higher token usage and computational cost~\cite{agentdiet,llmlingua}. \textbf{Second, degraded task performance.} In Figure~\ref{fig:intro} (right), identifying the name Sébastien Buemi suffices to guide the next search for his birth year, while extensive race details are irrelevant to this decision. Such redundant content consumes context and can distract the model from key evidence, impairing reasoning quality~\cite{lostinmiddle,longcontext2,longcontext3}.

A natural solution is to compress the agent's context. Existing methods mainly use token pruning or text summarization. Token pruning uses a model to identify and remove less informative tokens~\cite{SelectiveContext,llmlingua,llmlingua2,longllmlingua}. Text summarization uses an auxiliary LLM to condense tool outputs or interaction histories into shorter summaries~\cite{swe-pruner,recomp}. 
However, such text-level compression has a fundamental flaw: because the compressed result remains a discrete token sequence that must stay semantically coherent, its achievable compression ratio is inherently limited.

A closer look at tool–agent interactions raises a fundamental question: must tool outputs be expressed in textual tokens? Since these outputs primarily support an agent's subsequent reasoning and actions, tool–agent interaction need not rely on explicit textual tokens. This observation motivates us to reconsider how tool information is represented. Compared with discrete text tokens, continuous latent tokens (i.e., high-dimensional vectors in the embedding space) offer a more flexible alternative. These latent tokens can encode rich information compactly under a limited token budget, while their differentiability enables direct optimization for downstream objectives. Recent studies~\cite{latent_summary,latentlora} have explored latent tokens in reasoning~\cite{Coconut,CCoT,softcot,icot,latent_reasoning2}, agents' communication~\cite{kvcomm,latentmas,latentcomm2,agent_primitives}, and memory~\cite{memgen,nextmem,memoryllm,latentevolve}. Together, these studies suggest that models can directly process and exchange information through latent tokens, opening up new possibilities for information transfer between tools and agents.

Motivated by this analysis, we propose \name (\mbox{\textbf{Co}}\mbox{ntext} \textbf{C}om\-pression \mbox{vi\textbf{a}} \mbox{\textbf{Co}ntinuous} \mbox{\textbf{La}tent} Tokens), a latent context compression model for LLM agents. Conditioned on the task instruction and the agent's current-step reasoning trajectory, \name compresses long, noisy tool observations into a few latent tokens that retain information critical to subsequent decisions. These latent tokens are mapped into the agent's input embedding space, enabling it to use the compressed information for further reasoning and actions.

Although latent tokens enable efficient information compression, applying them to compress tool observations for agents introduces two key challenges. The first is to make compressed representations effectively usable by the agent. The second is to preserve decision-critical information under a limited representation budget. To address the first challenge, we design a \textbf{context-aware encoder with a latent adapter} that maps compressed representations into the target agent's input embedding space. This allows the agent to directly use these latent tokens for subsequent reasoning and actions. For the second challenge, we propose an \textbf{action-preserving training} method. Specifically, we align the agent's next-action distributions under full and compressed tool observations. This trains the compression model to minimize its impact on the agent's next-action behavior, thereby preserving action-critical information.

We evaluate \name across two model configurations and four web-search QA benchmarks. Compared to uncompressed agents, \name reduces total token consumption by 70.01\%--89.24\% and end-to-end runtime by 24.45\%--55.97\%, demonstrating substantial efficiency gains.
Despite aggressive compression, \name consistently outperforms the uncompressed agent in both EM and F1, and achieves the best EM and F1 among all methods in six of the eight model--dataset settings.

Our contributions are threefold:
\begin{itemize}
    \item To our knowledge, we are the first to recognize that tools need not convey information through explicit tokens and to compress tool outputs into latent tokens.

    \item We propose \name, a latent context compression model trained to preserve action-critical information.

    \item Experiments on web-search agents demonstrate that \name substantially improves efficiency while consistently improving task performance.
\end{itemize}

\section{Related Work}
\subsection{LLM-Based Agents}
\label{sec:2.1}

LLM-based agents typically follow the ReAct paradigm ~\cite{react}, alternating between reasoning and tool use to accomplish multi-step tasks. Formally, given a task instruction $q$, an agent parameterized by $\psi$ generates a reasoning step $r_t$ and issues a tool call $a_t$ at step $t$, receiving a tool observation $o_t$. Let $h_t=(r_1,a_1,o_1,\ldots,r_t,a_t)$ denote the trajectory prefix before receiving $o_t$, comprising previous reasoning steps, tool calls, and observations. The agent appends $o_t$ to its context and continues reasoning to select the next tool call or produce a final answer:
\begin{equation}
a_{t+1}\sim p_\psi(\cdot\mid q,h_t,o_t).
\end{equation}

As tool interactions proceed, the agent accumulates previous reasoning steps, tool calls, and their observations in its context. Without compression or truncation, the context length after the $t$-th tool call can be approximated as
\begin{equation}
K_t\approx |q|+\sum_{i=1}^{t}
\left(|r_i|+|a_i|+|o_i|\right).
\end{equation}
where $|\cdot|$ denotes the token count and $K_t$ is the total context length, excluding fixed prompt and formatting overhead. Since tool calls often return lengthy search results and document snippets, tool observations accumulate over successive interactions and can eventually dominate the agent's context, incurring substantial context overhead.

\subsection{Textual Compression for Tool-Augmented Agents}

To reduce the context overhead of tool interactions, existing methods primarily rely on token pruning and text summarization~\cite{swe-pruner,agentdiet,coact,longcodezip}. Formally, given the tool observation $o_t$ at step $t$, text-based compression methods transform it into a shorter sequence of discrete tokens using a compression function $g_\eta$:
\begin{equation}
\hat{o}_t = g_\eta(q,h_t,o_t) \in \mathcal{V}^{\ell_t},
\quad \ell_t < |o_t|,
\label{eq:text_compression}
\end{equation}
where $\mathcal{V}$ denotes the vocabulary, $\ell_t$ is the compressed sequence length. The agent then replaces the original observation $o_t$ with its compressed counterpart $\hat{o}_t$ to select the next action:
\begin{equation}
a_{t+1} \sim p_\psi(\cdot \mid q,h_t,\hat{o}_t).
\label{eq:text_action}
\end{equation}

 Token-pruning methods selectively remove less informative tokens. LLMLingua estimates token informativeness using language models~\citep{llmlingua}, while LLMLingua-2 formulates compression as a token classification task~\citep{llmlingua2}. LongLLMLingua further improves the density of task-relevant information in long-context prompts~\citep{longllmlingua}.

Text summarization instead employs auxiliary language models to condense tool outputs or interaction histories into shorter natural-language summaries. RECOMP generates query-relevant summaries of retrieved documents~\citep{recomp}, while Search-o1 extracts useful information from search results through its Reason-in-Documents module~\citep{search_o1,search-r1}. For long-horizon agents, ACON jointly compresses tool observations and interaction histories, refining compression instructions based on agent failures~\citep{acon}.

Despite their different strategies, these methods retain information in the form of discrete text tokens. In contrast, \name compresses tool observations into continuous latent representations, moving beyond text-based compression.

\subsection{Latent Tokens for Agents}

In recent years, continuous latent tokens have been explored for reasoning, communication, and memory in large language agents. For latent reasoning, Coconut~\citep{Coconut}, CCoT~\citep{CCoT}, and SoftCoT~\citep{softcot} explore reasoning in continuous latent spaces, reducing reliance on explicit natural-language reasoning processes. For agent communication, KVComm~\citep{kvcomm} and LatentMAS~\citep{latentmas} investigate information exchange between agents through hidden states or KV caches. For memory, MemGen~\citep{memgen} and NextMem~\citep{nextmem} explore storing and utilizing historical information through continuous latent tokens. These studies demonstrate that latent tokens can support both internal information processing and inter-model communication, offering new possibilities beyond traditional text-based interactions.

Another line of research employs latent tokens to compress long textual contexts~\cite{cocom,xrag,500xcompressor,autocompressor,gist,icae}. Gist Tokens~\citep{gist} learns special compression tokens to encode long prompts into compact representations; AutoCompressor~\citep{autocompressor} compresses long texts into continuous summary vectors; and 500xCompressor~\citep{500xcompressor} explores context compression at extremely high compression ratios. Among these approaches, ICAE~\citep{icae}, which is closely related to our work, employs an autoencoder-based architecture to compress long texts into a small number of continuous vectors and trains a language model to utilize these latent tokens for downstream tasks.

However, existing latent context compression methods primarily focus on preserving or reconstructing the original textual information. In contrast, \name{} directly optimizes for the agent's subsequent decisions, preserving decision-critical information while discarding redundant content.


\section{Method}
\label{sec:method}
\begin{figure*}[t]
    \centering
    \includegraphics[width=\textwidth]{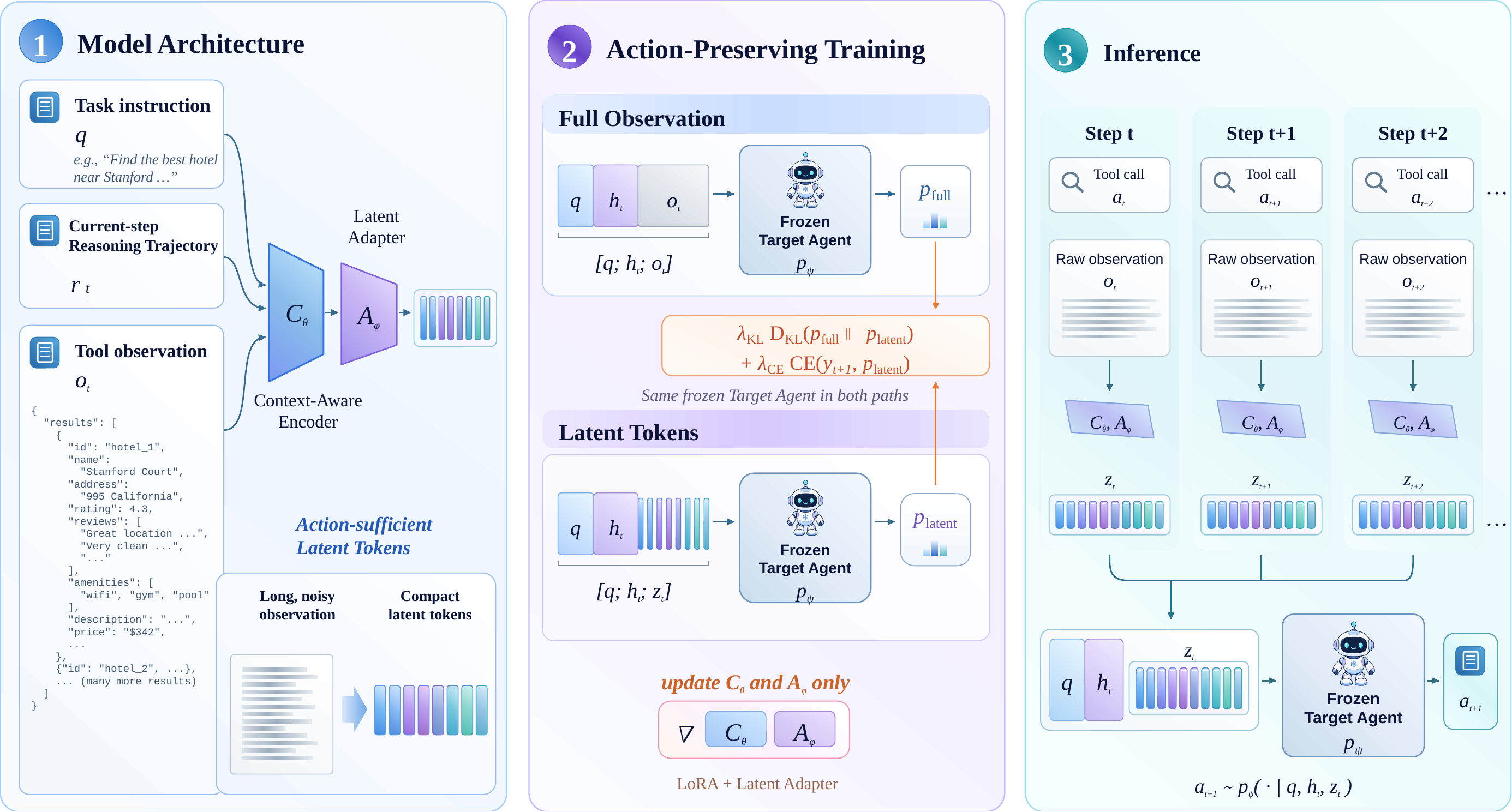}
    \caption{\textbf{Overview of \name.}
\textbf{Left:} The latent tool bottleneck encodes raw tool observations into compact continuous latent representations,
conditioned on the task instruction and current
reasoning state.
\textbf{Middle:} Action-preserving distillation aligns the next-action distributions of a full-observation teacher and a latent-conditioned student, updating only the compressor while keeping the agent backbone frozen.
\textbf{Right:} During inference, raw tool observations are replaced by latent representations after each tool call,
enabling efficient multi-step agent interactions.}
    \label{fig:overview}
\end{figure*}

\subsection{Overview}
As shown in Figure~\ref{fig:overview}, we propose \name{}, a continuous latent context compression model for LLM agents. Unlike traditional text-based compression methods, \name{} compresses lengthy tool observations into a small number of latent tokens to reduce the context overhead of tool interactions. We first theoretically analyze the information capacity of latent tokens, providing theoretical motivation for compressing tool observations into continuous vectors (Section ~\ref{sec:3.2}).

Specifically, after each tool call, \name{} takes the task instruction, the agent's current-step reasoning trajectory, and the tool observation as input. A lightweight encoder processes these inputs, while a latent adapter maps the resulting latent tokens into the target agent's input embedding space (Section ~\ref{sec:3.3}). To preserve decision-critical information under a limited representation budget, we further introduce \textbf{Action-Preserving Training}, which trains the compressor by aligning the agent's next-action distributions under full and compressed observations(Section ~\ref{sec:3.4}). During inference, the trained compressor replaces raw tool observations with latent tokens after each tool call, allowing the agent to perform multi-step interactions with a more compact context (Section ~\ref{sec:3.5}).

\subsection{Theoretical Analysis}
\label{sec:3.2}
We analyze the potential advantage of latent tokens through their finite-precision codebook capacity. Here, capacity is defined as the logarithm of the number of distinguishable codewords, with the context shared by the encoder and the agent held fixed.

For a model with vocabulary size $|\mathcal V|$, each text token is selected from $|\mathcal V|$ symbols. A sequence of $n$ tokens therefore admits at most $|\mathcal V|^n$ distinct sequences, giving a maximum codebook capacity of
\begin{equation}
    \mathcal C_{\mathrm{text}}(n)
    = n\log_2|\mathcal V|.
    \label{eq:text_codebook_capacity}
\end{equation}

A latent token can instead encode information jointly across multiple numerical coordinates within a single input position. Assume each $d$-dimensional latent token has $d_{\mathrm{eff}}\le d$ effective coordinates, each admitting $B$ distinguishable values at finite precision, and that all coordinate combinations across representations are realizable. Under these assumptions, $m$ representations have a codebook capacity of
\begin{equation}
    \mathcal C_{\mathrm{latent}}(m)
    = m d_{\mathrm{eff}}\log_2 B.
    \label{eq:latent_codebook_capacity}
\end{equation}

Thus, when $d_{\mathrm{eff}}\log_2 B>\log_2|\mathcal V|$, continuous latent representations can provide a larger codebook capacity under the same input-position budget, i.e., $m=n$. This potential comes from joint encoding across multiple coordinates, whereas each text position remains restricted to a symbol from a fixed vocabulary. This analysis motivates continuous compression of tool observations from a capacity perspective; the information usable in practice still depends on the compressor's encoding ability and the agent's ability to use the representations.

\subsection{Model Architecture}
\label{sec:3.3}

Building on the above theoretical analysis, we propose \name{}, a continuous latent context compression model for LLM agents. \name{} encodes tool observations into a small number of latent tokens and maps them into the target agent's input embedding space. To make these latent tokens directly usable by the agent, we design a compression model consisting of a \emph{context-aware encoder} and a \emph{latent adapter}, as illustrated in Figure~\ref{fig:overview} (left).

\noindent\textbf{Context-Aware Encoder.}
Our analysis of tool-use trajectories shows that a tool observation is not only relevant to the observation itself, but also depends on the task instruction and the agent's current reasoning state. To this end, we employ a small language model $C_\theta$ as the context-aware encoder, which jointly encodes the task instruction, current-step reasoning trajectory, and tool observation. Specifically, given the task instruction $q$, current-step reasoning trajectory prefix $r_t$, and tool observation $o_t$, the model produces context-conditioned hidden states:
\begin{equation}
H_t = C_\theta(q,r_t,o_t).
\label{eq:encoding}
\end{equation}

\noindent\textbf{Latent Adapter.}
Since the context-aware encoder and the target agent operate in different representation spaces, the encoded hidden states cannot be directly consumed by the agent. To bridge this gap, we introduce a latent adapter $A_\phi$ that transforms the hidden states into a fixed number of latent tokens and maps them into the target agent's input embedding space:
\begin{equation}
z_t = A_\phi(H_t) \in \mathbb{R}^{m\times d},
\label{eq:adapter}
\end{equation}
where $m$ is the number of latent tokens and $d$ is the target agent's input embedding dimension. The resulting latent tokens are then directly inserted into the agent's input embedding sequence, delimited by special boundary markers:
\begin{equation}
\begin{aligned}
X_t = \big[&
E_\psi(q,h_t);
E_\psi(\langle\mathrm{latent\_obs}\rangle);z_t;
E_\psi(\langle/\mathrm{latent\_obs}\rangle)
\big].
\end{aligned}
\label{eq:injection}
\end{equation}
Here, $E_\psi$ denotes the target agent's input embedding function. This design provides a continuous interface between tool observations and the target agent.

\subsection{Action-Preserving Training}
\label{sec:3.4}

Although latent tokens can encode tool observations compactly, preserving decision-critical information under a limited representation budget remains challenging. To address this challenge, we design a training method that aligns the agent's next-action distributions under full and compressed observations, encouraging the compression model to retain information critical to next-action selection.

Specifically, given the task instruction $q$, trajectory prefix $h_t$, and tool observation $o_t$, we feed the full observation $o_t$ and its compressed latent tokens $z_t$ separately into the same frozen target agent. The resulting next-action distributions are defined as
\begin{equation}
\begin{aligned}
p_{\mathrm{full}} &= p_\psi(\cdot\mid q,h_t,o_t),\\
p_{\mathrm{latent}} &= p_\psi(\cdot\mid q,h_t,z_t).
\end{aligned}
\label{eq:action_distributions}
\end{equation}
Our goal is to align these distributions, minimizing the impact of compression on the agent's subsequent decisions.

To this end, we jointly optimize an action-distribution alignment loss and an action-level supervision loss:
\begin{equation}
\begin{aligned}
\mathcal{L} ={} &
\lambda_{\mathrm{KL}}
D_{\mathrm{KL}}(p_{\mathrm{full}}\parallel p_{\mathrm{latent}}) + \lambda_{\mathrm{CE}}
\mathrm{CE}(y_{t+1},p_{\mathrm{latent}}),
\end{aligned}
\label{eq:training_objective}
\end{equation}
where $y_{t+1}$ denotes the next-action token recorded in the collected trajectory, and $\lambda_{\mathrm{KL}}$ and $\lambda_{\mathrm{CE}}$ control the respective loss weights. The KL divergence aligns the action distributions under full and compressed observations, while the cross-entropy loss further supervises the predictions using recorded actions. Together, these objectives encourage the compression model to prioritize information relevant to subsequent decisions.

Since latent tokens are continuous and differentiable, gradients can propagate through the frozen target agent to optimize the compression model:
\begin{equation}
\nabla_{\theta,\phi}\mathcal{L}
=
\left(
\frac{\partial z_t}{\partial(\theta,\phi)}
\right)^{\!\top}
\nabla_{z_t}\mathcal{L}.
\label{eq:compression_gradient}
\end{equation}
During training, we keep the target agent frozen and update only the LoRA parameters of the context-aware encoder and the parameters of the latent adapter. By directly aligning the agent's action distributions under full and compressed observations, \name{} learns action-sufficient latent tokens without reconstructing the original tool observations.

\subsection{Inference}
\label{sec:3.5}

At inference, \name replaces each raw tool observation with latent tokens. Given the task instruction $q$, current-step trajectory prefix $r_t$, and tool observation $o_t$, the trained compression model produces
\begin{equation}
z_t = A_\phi\!\left(C_\theta(q,r_t,o_t)\right).
\label{eq:inference_compression}
\end{equation}
The agent then conditions on $z_t$, rather than the raw observation, to generate its next action:
\begin{equation}
a_{t+1}\sim p_\psi(\cdot\mid q,h_t,z_t).
\label{eq:inference_action}
\end{equation}
For multi-step tool use, \name repeats this process after every tool call. By replacing tool outputs with a small number of latent tokens, \name reduces context growth while retaining information needed for subsequent decisions.

\section{Experimental Setup}
\label{sec:experimental_setup}

\begin{table*}[t]
\centering
\caption{
\textbf{Efficiency comparison} across four web-search
QA datasets using Qwen3.5-35B-A3B and Qwen3.5-4B.
Best results are in bold. Blue annotations under
\name\ indicate reductions (\%) relative to Vanilla
in token usage and runtime.
}
\label{tab:efficiency_qwen}

\footnotesize
\setlength{\tabcolsep}{2.5pt}
\renewcommand{\arraystretch}{1.25}

\begin{tabular*}{\textwidth}{
@{\extracolsep{\fill}}
lcccc>{\columncolor{EffColaTint}}c
@{\hspace{1.3em}}
cccc>{\columncolor{EffColaTint}}c@{}
}

\toprule[0.85pt]


& \multicolumn{5}{
>{\columncolor{EffHeaderBlue}}c
}{
\strut\textbf{\color{EffInkBlue}NQ}
}
& \multicolumn{5}{
>{\columncolor{EffHeaderBlue}}c
}{
\strut\textbf{\color{EffInkBlue}TriviaQA}
} \\

\cmidrule(lr){2-6}
\cmidrule(l){7-11}

& Vanilla
& \makecell[c]{LongLLM\\Lingua}
& \makecell[c]{LLM\\Summary}
& ICAE
& {\color{EffInkBlue}\name}
& Vanilla
& \makecell[c]{LongLLM\\Lingua}
& \makecell[c]{LLM\\Summary}
& ICAE
& {\color{EffInkBlue}\name} \\

\midrule[0.55pt]

$I_{\mathrm{token}}\,\downarrow$
& 14362.12
& 8607.04
& 6099.83
& 5926.23
& \effBestDelta{1766.53}{87.7}
& 10022.93
& 5445.57
& 3791.53
& 3038.35
& \effBestDelta{1413.12}{85.9} \\

\addlinespace[3pt]

$O_{\mathrm{token}}\,\downarrow$
& 992.69
& 1232.61
& 1080.51
& 665.57
& \effBestDelta{417.69}{57.9}
& 779.34
& 1048.02
& 679.36
& 535.42
& \effBestDelta{349.36}{55.2} \\

\addlinespace[3pt]

$\mathrm{Comp\_rate}\,\uparrow$
& --
& 35.92\%
& 53.24\%
& 57.07\%
& \textbf{85.78\%}
& --
& 39.89\%
& 58.61\%
& 66.92\%
& \textbf{83.68\%} \\

\addlinespace[3pt]

\textit{Time (s)}\,$\downarrow$
& 96.19
& 109.48
& 121.02
& 87.06
& \effBestDelta{64.64}{32.8}
& 73.30
& 86.91
& 91.67
& 64.89
& \effBestDelta{55.38}{24.4} \\

\midrule[0.9pt]


& \multicolumn{5}{
>{\columncolor{EffHeaderBlue}}c
}{
\strut\textbf{\color{EffInkBlue}2Wiki}
}
& \multicolumn{5}{
>{\columncolor{EffHeaderBlue}}c
}{
\strut\textbf{\color{EffInkBlue}HotpotQA}
} \\

\cmidrule(lr){2-6}
\cmidrule(l){7-11}

& Vanilla
& \makecell[c]{LongLLM\\Lingua}
& \makecell[c]{LLM\\Summary}
& ICAE
& {\color{EffInkBlue}\name}
& Vanilla
& \makecell[c]{LongLLM\\Lingua}
& \makecell[c]{LLM\\Summary}
& ICAE
& {\color{EffInkBlue}\name} \\

\midrule[0.55pt]

$I_{\mathrm{token}}\,\downarrow$
& 37427.21
& 21110.06
& \textbf{10011.31}
& 12268.64
& \effDelta{10744.51}{71.3}
& 42723.75
& 21241.95
& \textbf{8088.21}
& 15763.53
& \effDelta{9429.04}{77.9} \\

\addlinespace[3pt]

$O_{\mathrm{token}}\,\downarrow$
& 1510.37
& 1488.63
& 1163.11
& 1080.53
& \effBestDelta{932.57}{38.3}
& 1560.28
& 1596.04
& 1253.70
& 1035.63
& \effBestDelta{831.43}{46.7} \\

\addlinespace[3pt]

$\mathrm{Comp\_rate}\,\uparrow$
& --
& 41.96\%
& \textbf{71.30\%}
& 65.72\%
& 70.01\%
& --
& 48.43\%
& \textbf{78.90\%}
& 62.06\%
& 76.83\% \\

\addlinespace[3pt]

\textit{Time (s)}\,$\downarrow$
& 164.99
& 137.22
& 175.97
& 142.18
& \effBestDelta{119.91}{27.3}
& 161.39
& 145.83
& 160.24
& 139.04
& \effBestDelta{120.88}{25.1} \\

\bottomrule[0.85pt]

\end{tabular*}
\end{table*}
\begin{table*}[t]
\centering
\caption{
\textbf{Efficiency comparison} across four web-search
QA datasets using GLM-4.7-Flash and GLM-Z1-9B.
Best results are in bold. Blue annotations under
\name\ indicate reductions (\%) relative to Vanilla
in token usage and runtime.
}
\label{tab:efficiency_glm}

\footnotesize
\setlength{\tabcolsep}{2.5pt}
\renewcommand{\arraystretch}{1.25}

\begin{tabular*}{\textwidth}{
@{\extracolsep{\fill}}
lcccc>{\columncolor{EffColaTint}}c
@{\hspace{1.3em}}
cccc>{\columncolor{EffColaTint}}c@{}
}

\toprule[0.85pt]


& \multicolumn{5}{
>{\columncolor{EffHeaderBlue}}c
}{
\strut\textbf{\color{EffInkBlue}NQ}
}
& \multicolumn{5}{
>{\columncolor{EffHeaderBlue}}c
}{
\strut\textbf{\color{EffInkBlue}TriviaQA}
} \\

\cmidrule(lr){2-6}
\cmidrule(l){7-11}

& Vanilla
& \makecell[c]{LongLLM\\Lingua}
& \makecell[c]{LLM\\Summary}
& ICAE
& {\color{EffInkBlue}\name}
& Vanilla
& \makecell[c]{LongLLM\\Lingua}
& \makecell[c]{LLM\\Summary}
& ICAE
& {\color{EffInkBlue}\name} \\

\midrule[0.55pt]

$I_{\mathrm{token}}\,\downarrow$
& 17388.56
& 5779.61
& \textbf{3903.19}
& 5068.65
& \effDelta{4265.42}{75.5}
& 11176.91
& 4689.20
& 2842.51
& 3510.28
& \effBestDelta{2595.00}{76.8} \\

\addlinespace[3pt]

$O_{\mathrm{token}}\,\downarrow$
& 887.74
& 797.81
& 734.77
& 805.93
& \effBestDelta{536.37}{39.6}
& 738.96
& 688.27
& 634.84
& 617.94
& \effBestDelta{413.28}{44.1} \\

\addlinespace[3pt]

$\mathrm{Comp\_rate}\,\uparrow$
& --
& 64.01\%
& \textbf{74.62\%}
& 67.86\%
& 73.73\%
& --
& 54.87\%
& 70.82\%
& 65.36\%
& \textbf{74.75\%} \\

\addlinespace[3pt]

\textit{Time (s)}\,$\downarrow$
& 89.29
& 91.06
& 106.47
& 83.63
& \effBestDelta{42.17}{52.8}
& 91.36
& 89.93
& 128.36
& 72.33
& \effBestDelta{41.97}{54.1} \\

\midrule[0.9pt]


& \multicolumn{5}{
>{\columncolor{EffHeaderBlue}}c
}{
\strut\textbf{\color{EffInkBlue}2Wiki}
}
& \multicolumn{5}{
>{\columncolor{EffHeaderBlue}}c
}{
\strut\textbf{\color{EffInkBlue}HotpotQA}
} \\

\cmidrule(lr){2-6}
\cmidrule(l){7-11}

& Vanilla
& \makecell[c]{LongLLM\\Lingua}
& \makecell[c]{LLM\\Summary}
& ICAE
& {\color{EffInkBlue}\name}
& Vanilla
& \makecell[c]{LongLLM\\Lingua}
& \makecell[c]{LLM\\Summary}
& ICAE
& {\color{EffInkBlue}\name} \\

\midrule[0.55pt]

$I_{\mathrm{token}}\,\downarrow$
& 31693.47
& 11934.39
& 5113.19
& 7677.77
& \effBestDelta{3003.38}{90.5}
& 28781.88
& 6059.07
& 4690.11
& \textbf{4029.96}
& \effDelta{4041.03}{86.0} \\

\addlinespace[3pt]

$O_{\mathrm{token}}\,\downarrow$
& 976.35
& 672.11
& 895.04
& 777.21
& \effBestDelta{511.07}{47.7}
& 834.04
& 826.07
& 747.77
& 707.66
& \effBestDelta{609.95}{26.9} \\

\addlinespace[3pt]

$\mathrm{Comp\_rate}\,\uparrow$
& --
& 61.41\%
& 81.61\%
& 74.12\%
& \textbf{89.24\%}
& --
& 76.75\%
& 81.64\%
& 84.00\%
& \textbf{84.30\%} \\

\addlinespace[3pt]

\textit{Time (s)}\,$\downarrow$
& 106.37
& 109.93
& 153.25
& 89.21
& \effBestDelta{46.83}{56.0}
& 84.55
& 102.54
& 101.61
& 99.58
& \effBestDelta{49.37}{41.6} \\

\bottomrule[0.85pt]

\end{tabular*}
\end{table*}
\begin{table*}[t]
\centering
\caption{
\textbf{Web search performance comparison.}
Best results within each model setting are in bold.
Arrows in the \name\ rows indicate relative changes
(\%) over the Vanilla Model for each metric.
}
\label{tab:performance}

\footnotesize
\setlength{\tabcolsep}{4pt}
\renewcommand{\arraystretch}{1.12}

\begin{tabularx}{\textwidth}{
@{}l*{8}{>{\centering\arraybackslash}X}@{}
}

\toprule[0.9pt]

\multirow{3}{*}{\textbf{Method}}
& \multicolumn{4}{
>{\columncolor{HeaderBlue}}c
}{
\strut\textbf{\color{ColaBlue}Single-hop QA}
}
& \multicolumn{4}{
>{\columncolor{HeaderBlue}}c
}{
\strut\textbf{\color{ColaBlue}Multi-hop QA}
} \\

\cmidrule(lr){2-5}
\cmidrule(l){6-9}

& \multicolumn{2}{c}{\textbf{NQ}}
& \multicolumn{2}{c}{\textbf{TriviaQA}}
& \multicolumn{2}{c}{\textbf{2Wiki}}
& \multicolumn{2}{c}{\textbf{HotpotQA}} \\

\cmidrule(lr){2-3}
\cmidrule(lr){4-5}
\cmidrule(lr){6-7}
\cmidrule(l){8-9}

& EM & F1
& EM & F1
& EM & F1
& EM & F1 \\

\midrule[0.75pt]


\rowcolor{GroupGray}
\multicolumn{9}{@{}c@{}}{
\rule{0pt}{2.65ex}
\textbf{Qwen3.5-35B-A3B \& Qwen3.5-4B}
} \\

Vanilla Model
& 23.67 & 33.17
& 56.10 & 62.87
& 17.30 & 26.94
& 11.00 & 16.41 \\

LongLLMLingua
& 19.00 & 25.81
& 44.30 & 48.69
& 21.30 & \textbf{34.25}
& 11.90 & 19.75 \\

LLMSummary
& 24.65 & 32.35
& 54.70 & 61.38
& 21.06 & 31.77
& 23.10 & 29.60 \\

ICAE
& 24.38 & 35.20
& \textbf{61.80} & \textbf{69.27}
& 13.72 & 16.91
& 23.71 & 31.21 \\

\rowcolor{ColaTint}
\textbf{\name}
& \colaBestGain{24.78}{4.69}
& \colaBestGain{37.33}{12.54}
& \makecell[c]{59.20\\[-1pt]\colaGainUp{5.53}}
& \makecell[c]{67.07\\[-1pt]\colaGainUp{6.68}}
& \colaBestGain{24.00}{38.73}
& \makecell[c]{32.34\\[-1pt]\colaGainUp{20.04}}
& \colaBestGain{25.70}{133.64}
& \colaBestGain{33.75}{105.67} \\

\midrule[0.75pt]


\rowcolor{GroupGray}
\multicolumn{9}{@{}c@{}}{
\rule{0pt}{2.65ex}
\textbf{GLM-4.7-Flash \& GLM-Z1-9B}
} \\

Vanilla Model
& 21.50 & 26.08
& 41.50 & 44.57
& 26.50 & 31.29
& 20.50 & 25.89 \\

LongLLMLingua
& 17.00 & 21.16
& 39.50 & 42.41
& 20.50 & 24.91
& 19.50 & 23.00 \\

LLMSummary
& 16.00 & 21.05
& 46.00 & 49.29
& 22.50 & 25.61
& 13.50 & 17.60 \\

ICAE
& 17.00 & 20.82
& 38.00 & 40.94
& 18.50 & 20.75
& 16.50 & 20.13 \\

\rowcolor{ColaTint}
\textbf{\name}
& \colaBestGain{23.50}{9.30}
& \colaBestGain{31.45}{20.59}
& \colaBestGain{51.50}{24.10}
& \colaBestGain{59.41}{33.30}
& \colaBestGain{28.00}{5.66}
& \colaBestGain{36.20}{15.69}
& \colaBestGain{23.50}{14.63}
& \colaBestGain{30.76}{18.81} \\

\bottomrule[0.9pt]

\end{tabularx}
\end{table*}

\subsection{Research Questions}
\label{sec:research_questions}

We systematically evaluate \name{} to address the following four research questions:

\noindent\textbf{RQ1 (Efficiency): How much computational overhead can \name{} reduce in terms of token consumption and execution time?}

We measure token consumption, compression rates, and end-to-end runtime across different model configurations and benchmarks.

\noindent\textbf{RQ2 (Task Performance): Can \name{} preserve or improve agent task performance while reducing context overhead?}

We compare EM and F1 scores against uncompressed agents and existing compression methods to assess the impact of compression on task performance.

\noindent\textbf{RQ3 (Ablation Study): What factors affect the performance of \name{}?}

We investigate how the number of latent tokens and different contextual inputs affect compression effectiveness and task performance.

\noindent\textbf{RQ4 (Case Study): How does \name{} support agent decision-making through latent tokens?}

We present a representative case study to illustrate how latent tokens preserve decision-critical information and guide subsequent actions.

\subsection{Agent Setting and Model Configurations}
\label{sec:agent_setting}

We evaluate \name in a web-search agent setting, as web search is one of the most common tool-use scenarios for LLM agents~\cite{search_o1, search-r1}, and its returned results are typically lengthy and highly redundant (Figure~\ref{fig:intro}), making it a natural testbed for observation compression. Notably, \name involves no search-specific design: the compressor conditions only on the task instruction, the current reasoning trajectory, and the raw tool observation, and can thus be applied to other tools without architectural modification.

To evaluate \name{} across different model configurations, we use two model pairs: \textbf{Qwen3.5-35B-A3B with Qwen3.5-4B}~\cite{qwen35}, and \textbf{GLM-4.7-Flash with GLM-Z1-9B}~\cite{glm,chatglm}.

\subsection{Baselines}
\label{sec:baselines}

We compare \name{} against four representative approaches:

\noindent\textbf{Vanilla Model.}
An uncompressed agent that directly appends raw tool observations to its context.

\noindent\textbf{LongLLMLingua.}
A token-pruning method that compresses context by selectively removing less informative tokens~\cite{longllmlingua}.

\noindent\textbf{LLMSummary.} 
A text summarization approach that uses an auxiliary language model to summarize tool observations into shorter natural-language text ~\cite{search_o1}.

\noindent\textbf{ICAE.} 
ICAE~\cite{icae} learns continuous latent tokens through text reconstruction. Since ICAE was not originally designed for agent tool observations, we adapt its training objective to our setting. Specifically, we use the same compression architecture as \name{} but replace the action-preserving training objective with a text-reconstruction objective, training the model to recover the original tool observations from compressed latent tokens.

\subsection{Benchmarks}
\label{sec:benchmarks}

We evaluate on four open-domain question answering benchmarks: \textbf{Natural Questions (NQ)}~\cite{nq}, \textbf{TriviaQA}~\cite{triviaqa}, \textbf{2WikiMultiHopQA (2Wiki)}~\cite{2wiki}, and \textbf{HotpotQA}~\cite{hotpotqa}.

NQ and TriviaQA primarily assess the agent's ability to retrieve and utilize factual evidence, while 2Wiki and HotpotQA require multi-hop reasoning across multiple pieces of evidence. Together, these benchmarks evaluate whether latent compression preserves critical information under varying reasoning complexity.

\subsection{Evaluation Metrics}
\label{sec:metrics}

We evaluate each method in terms of \textbf{task performance} and \textbf{system efficiency}.

\noindent\textbf{Task Performance.}
We report Exact Match (EM) and token-level F1 scores on the final answers to measure answer correctness.

\noindent\textbf{System Efficiency.}
We measure the average input tokens $I_{\mathrm{token}}$, output tokens $O_{\mathrm{token}}$, compression ratio $\mathrm{Comp\_rate}$, and end-to-end execution time \textit{Time}.

The compression ratio is defined as
\begin{equation}
\mathrm{Comp\_rate}
= 1-
\frac{
I_{\mathrm{token}}^{\mathrm{latent}}
+
O_{\mathrm{token}}^{\mathrm{latent}}
}{
I_{\mathrm{token}}^{\mathrm{raw}}
+
O_{\mathrm{token}}^{\mathrm{raw}}
},
\label{eq:compression_ratio}
\end{equation}
where the superscripts $\mathrm{latent}$ and $\mathrm{raw}$ denote the compressed and uncompressed settings, respectively.

\section{Experimental Results}
\label{sec:experiments}
\subsection{RQ1: How much computational overhead can \name{} reduce?}

In this RQ, we evaluate the efficiency gains of \name{} in terms of token consumption and end-to-end execution time across different model configurations and QA benchmarks. The results are presented in Tables~\ref{tab:efficiency_qwen} and~\ref{tab:efficiency_glm}.

\noindent\textbf{\name{} substantially reduces tool-context overhead.}
Compared with the Vanilla Model, which directly incorporates full tool observations into the agent's context, \name{} consistently reduces both input and output token consumption across all evaluated datasets. For instance, on NQ with the Qwen models, the average number of input tokens decreases from 14,362.12 to 1,766.53, representing an 87.70\% reduction. Similarly, on 2Wiki with the GLM models, input token consumption drops from 31,693.47 to 3,003.38, corresponding to a 90.52\% reduction. These results demonstrate that replacing noisy tool observations with compact latent tokens effectively mitigates context growth while reducing subsequent generation overhead.

\noindent\textbf{\name{} achieves high compression rates across all settings.}
\name{} reduces total token consumption by 70.01\%--89.24\% and achieves the highest compression rate in five of the eight model--dataset settings. Compared with ICAE, which shares the same architecture but uses a reconstruction objective, \name{} achieves a higher compression rate in all eight settings; for example, on 2Wiki with the GLM models, it reaches 89.24\% versus 74.12\% for ICAE. LLMSummary attains slightly higher compression rates in the remaining three settings, but with lower EM and F1 than \name{} in all three; for instance, on NQ with the GLM models, its EM is only 16.00\%, compared with 23.50\% for \name{} (Table~\ref{tab:performance}).

\noindent\textbf{\name{} translates token savings into tangible improvements in inference efficiency.}
Across all four datasets and both model configurations, \name{} consistently achieves the lowest average execution time. With the Qwen models, runtime decreases by 24.45\%--32.80\%; for example, on NQ it drops from 96.19 to 64.64 seconds, a 32.80\% reduction. The gains are even larger with the GLM models, where runtime decreases by 41.61\%--55.97\%; for example, on 2Wiki it drops from 106.37 to 46.83 seconds, a 55.97\% reduction. These findings demonstrate that the benefits of \name{} extend beyond token-level compression, translating into practical improvements in inference latency and computational cost.

\begin{figure}[t]
    \centering
    \includegraphics[width=0.6\columnwidth]{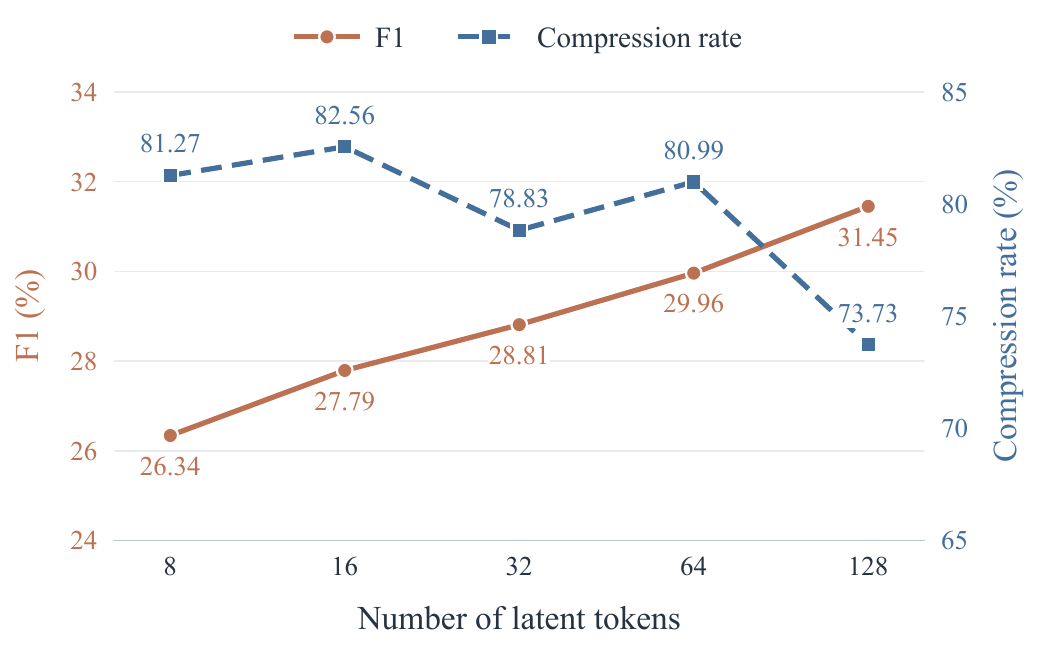}
      \caption{Effect of the number of latent tokens $m$ on NQ (GLM-4.7-Flash \& GLM-Z1-9B). F1 improves steadily as the latent budget increases.}
    \label{fig:num_latent}
\end{figure}

\subsection{RQ2: Can \name improve agent task performance while reducing tool-token overhead?}

In this RQ, we evaluate whether \name{} can preserve or improve agent performance while reducing tool-context overhead, using Exact Match (EM) and F1. The results are presented in Table~\ref{tab:performance}.

\noindent\textbf{High compression rates do not compromise task performance, but can even lead to improvements.}
Compared with the Vanilla Model, \name{} consistently achieves higher EM and F1 across all evaluated settings. On TriviaQA with Qwen, \name{} improves F1 from 62.87\% to 67.07\% while achieving an 83.68\% compression rate. Similarly, on 2Wiki with GLM, F1 increases from 31.29\% to 36.20\% at an 89.24\% compression rate. These results suggest that filtering redundant and irrelevant tool information may alleviate attention dilution caused by lengthy contexts, allowing agents to focus more effectively on task-relevant evidence.

\noindent\textbf{\name{} outperforms existing compression methods in most settings.}
\name{} achieves the highest score in 13 of the 16 dataset--metric pairs, and the highest EM and F1 simultaneously in six of the eight model--dataset combinations, including all four with the GLM models. For example, on TriviaQA with GLM, \name{} achieves an F1 of 59.41\%, surpassing the strongest baseline, LLMSummary (49.29\%). Similarly, on HotpotQA with Qwen, \name{} achieves an F1 of 33.75\%, compared with 31.21\% for the strongest baseline, ICAE. The only exceptions occur with the Qwen models, where ICAE achieves higher EM and F1 on TriviaQA and LongLLMLingua achieves a higher F1 on 2Wiki; however, these baselines compress tool observations considerably less than \name{} (66.92\% vs.\ 83.68\% and 41.96\% vs.\ 70.01\%, respectively). Unlike text-based or generic latent compression, \name{} prioritizes decision-relevant information, even under high compression rates.

\subsection{RQ3: What factors affect the performance of \name{}?}

\paragraph{\noindent\textbf{The number of latent tokens.}}
On NQ with the GLM models, we vary the number of latent tokens $m \in \{8, 16, 32, 64, 128\}$ and train a separate compressor for each setting (Figure~\ref{fig:num_latent}). F1 increases steadily from 26.34\% to 31.45\% as $m$ grows, while the compression rate stays between 78.83\% and 82.56\% for $m \le 64$ and drops to 73.73\% at $m=128$. We adopt $m=128$ as the default, as it offers the best trade-off between efficiency and task performance across datasets in our preliminary experiments.

\paragraph{\noindent\textbf{Different contextual inputs.}}
To assess the contribution of the conditioning signals, we train four variants with $m=128$: \textit{Full} $(q, r_t, o_t)$, \textit{w/o Reasoning State} $(q, o_t)$, \textit{w/o Task Goal} $(r_t, o_t)$, and \textit{Observation-only} $(o_t)$, where the task goal text is also removed from $r_t$ when excluded (Table~\ref{tab:context_ablation}). Both signals contribute: removing the reasoning state, the task goal, and both reduces EM by 2.05, 3.48, and 5.16 points, respectively. This confirms that decision-critical information cannot be fully determined from the observation alone. Removing the reasoning state causes a smaller drop, likely because on single-hop NQ the agent's current sub-goal largely coincides with the original question. Although the ablated variants achieve higher compression rates (78.38\%--80.43\%), they do so at the cost of EM. Notably, even the Observation-only variant outperforms all compression baselines on NQ with the GLM models in both EM (18.34\% vs.\ at most 17.00\%) and compression rate (79.94\% vs.\ at most 74.62\%), including ICAE, which shares the same architecture but uses a reconstruction objective. This highlights the benefit of action-preserving training.

\begin{table}[t]
\centering
\caption{\textbf{Context-component ablation} on NQ (GLM-4.7-Flash \& GLM-Z1-9B, $m=128$).}
\label{tab:context_ablation}

\normalsize
\setlength{\tabcolsep}{4pt}
\renewcommand{\arraystretch}{1.2}

\begin{tabularx}{\columnwidth}{
    @{}>{\raggedright\arraybackslash}Xcc@{}
}
\toprule[0.9pt]

\textbf{Method}
& \textbf{EM (\%)}
& \textbf{Compression rate (\%)} \\

\midrule[0.75pt]

\rowcolor{GroupGray}
\multicolumn{3}{@{}c@{}}{
    \rule{0pt}{2.65ex}\textbf{NQ}
} \\
\midrule

\rowcolor{ColaTint}
\textbf{Full}
& \textbf{23.50}
& 73.73 \\

w/o Reasoning State
& 21.45
& \textbf{80.43} \\

w/o Task Goal
& 20.02
& 78.38 \\

Observation-only
& 18.34
& 79.94 \\

\bottomrule[0.9pt]
\end{tabularx}
\end{table}

\begin{figure}[t]
    \centering
    \includegraphics[width=\columnwidth]{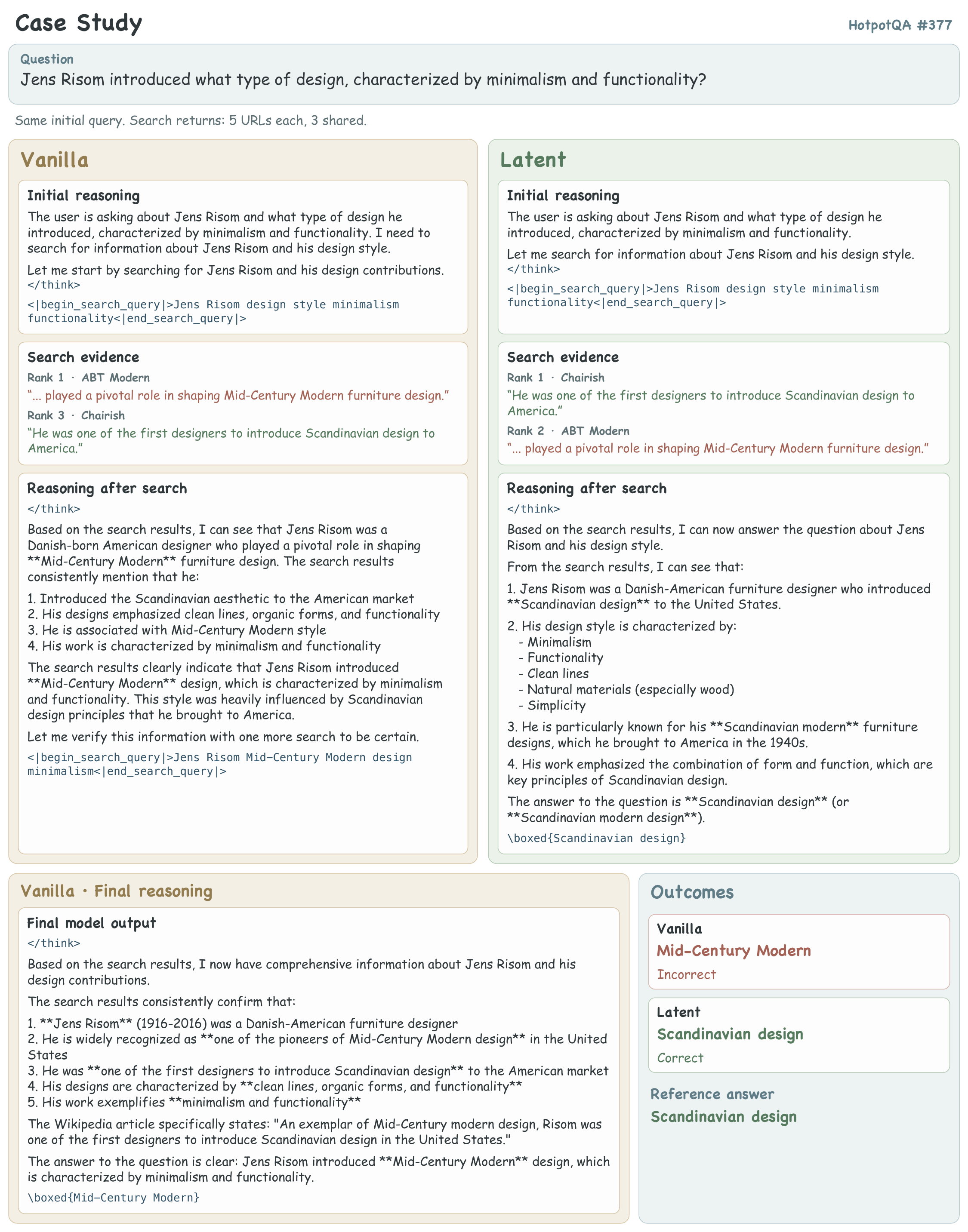}
    \caption{Case study on HotpotQA \#377 comparing the Vanilla Model and \name{}.}
    \label{fig:case_study}
\end{figure}

\subsection{RQ4: How does \name{} support agent decision-making through latent tokens?}

Figure~\ref{fig:case_study} presents a representative case from HotpotQA. The search results contain both information linking Jens Risom to Mid-Century Modern design and key evidence that he introduced Scandinavian design to America. The Vanilla Model fails to distinguish between these two pieces of information and produces an incorrect answer even after an additional search. In contrast, \name{} identifies the evidence directly relevant to the question and arrives at the correct answer with fewer searches and lower context overhead.

\section{Conclusion}
\label{sec:conclusion}

In this paper, we introduce \name, a continuous latent context compression model for compacting noisy tool observations in LLM agents. Unlike conventional token pruning and summarization methods, \name encodes tool outputs into a small number of task-conditioned latent tokens that are directly injected into the agent.
Through our action-preserving training, we learn a compression model that preserves decision-critical information rather than reconstructing the original observations.
Experiments on web-search tasks demonstrate that \name substantially reduces context overhead while consistently improving task performance.

More fundamentally, we challenge a prevailing assumption in LLM agents: interaction between tools and agents need not rely on explicit text tokens.
Instead, tools only need to convey information sufficient for subsequent decisions through compact, latent tokens. This perspective opens a promising direction for building more efficient and scalable long-horizon agents.

\bibliographystyle{preprint}
\bibliography{references}

\clearpage
\appendix
\section{Implementation Details}
\label{app:details}

\subsection{Training Configuration}

We construct the compression dataset from 4,188 distinct tasks,
containing 4,363 tool observation--action pairs.
We reserve 10\% of the pairs for validation, yielding
3,927 training pairs and 436 validation pairs.

The target agent remains frozen throughout training.
We update only the LoRA parameters of the context-aware model and the parameters of the latent adapter.
The LoRA rank, scaling factor, and dropout rate are set to 16, 32, and 0.05, respectively.

We use paged 8-bit AdamW as the optimizer, with initial learning rates of $1.5 \times 10^{-4}$ for both the LoRA parameters and the latent adapter.
The learning rate decays linearly to $5 \times 10^{-5}$. The weight decay is set to 0.
The loss weights $\lambda_{\mathrm{KL}}$ and $\lambda_{\mathrm{CE}}$ are set to 0.3 and 1.0, respectively. The KL loss weight is warmed up over the first 20\% of training steps.

\subsection{Inference Configuration}

During inference, each tool observation is compressed into 128 continuous latent tokens and directly injected into the target agent's input embedding sequence.

For web search, we retrieve up to 5 results per tool call and allow a maximum of 5 tool calls per task.

\subsection{Hardware and Software}

All experiments are conducted on 4 NVIDIA A100-SXM4 GPUs, each with 80\,GB of memory.

\paragraph{Training.}
For \name, the frozen target agent is distributed across 3 GPUs, while the context-aware encoder and the latent adapter are placed on the remaining GPU. 

\paragraph{Inference.}
All inference results, including the runtimes reported in Tables~\ref{tab:efficiency_qwen} and~\ref{tab:efficiency_glm}, are obtained with the target agent running on a single GPU in all settings. Where an additional GPU is used, it hosts only the compression model.

\section{Web Search Agent Prompt}
\label{app:search_prompt}

\begin{figure}[t]
    \centering
    \includegraphics[width=\columnwidth]{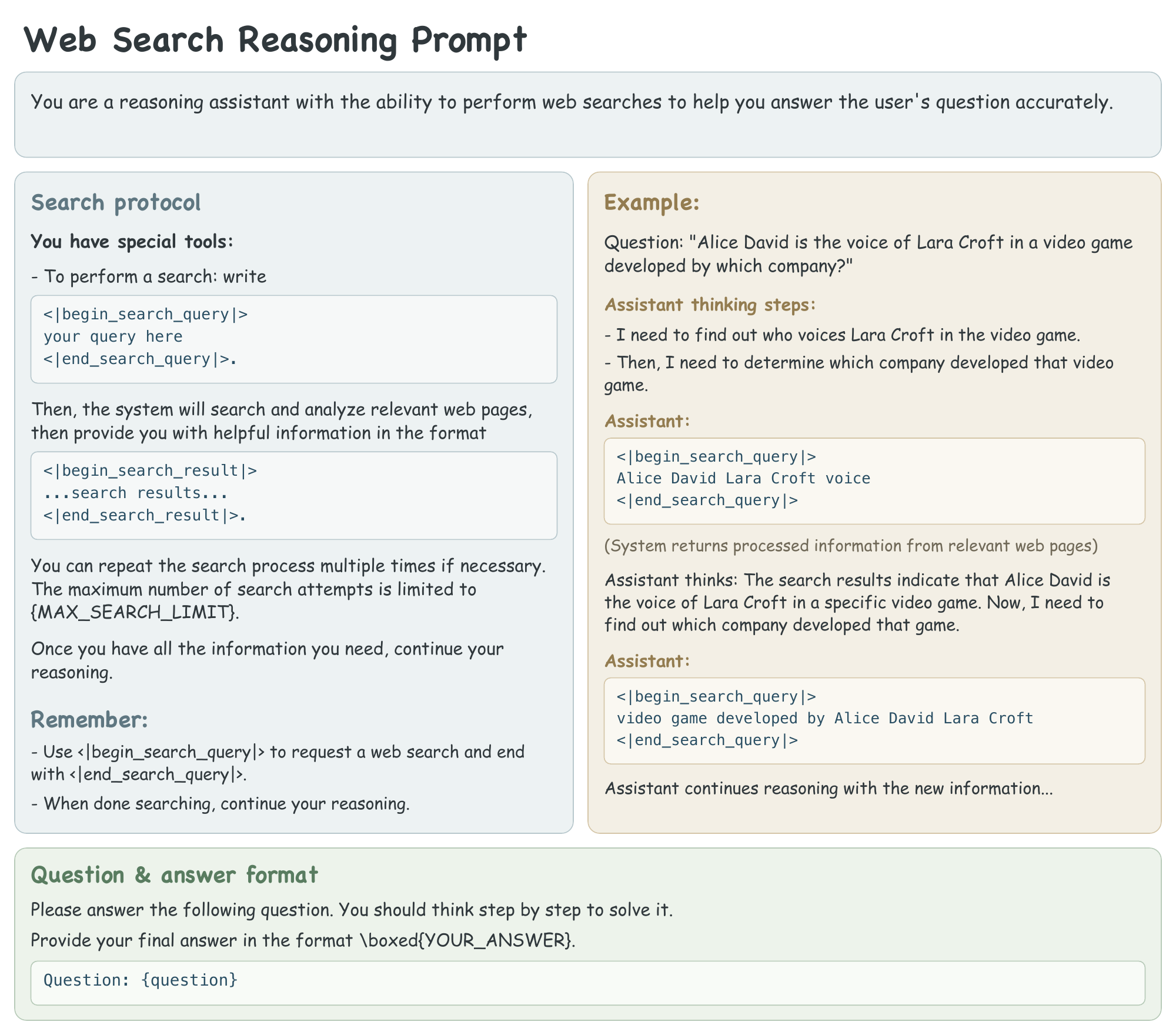}
    \caption{Web-search reasoning prompt used in our experiments.}
    \label{fig:web_search_prompt}
\end{figure}

Figure~\ref{fig:web_search_prompt} presents the prompt used to guide the agent in web-search-based question answering. The prompt specifies the search tool interface, the maximum number of search attempts, and the required final-answer format. It also provides an example of multi-step search to illustrate how the agent can iteratively retrieve information and reason over search results.

\end{document}